# Toward an AI Reasoning-Enabled System for Patient-Clinical Trial Matching

Caroline N. Leach, BS[1], Mitchell A. Klusty, BS[1], Samuel E. Armstrong, MS[1], Justine C. Pickarski, MS LGC[2], Kristen L. Hankins, MA[1], Emily B. Collier, MS[1], Maya Shah[1], Aaron D. Mullen, MS[1], V. K. Cody Bumgardner, PhD[1]

[1]Center For Applied Artificial Intelligence, University of Kentucky, Lexington, KY
[2]Molecular Tumor Board, Markey Cancer Center, Lexington, KY

**Abstract**

*Screening patients for clinical trial eligibility remains a manual, time-consuming, and resource-intensive process. We present a secure, scalable proof-of-concept system for Artificial Intelligence (AI)-augmented patient-trial matching that addresses key implementation challenges: integrating heterogeneous electronic health record (EHR) data, facilitating expert review, and maintaining rigorous security standards. Leveraging open-source, reasoning-enabled large language models (LLMs), the system moves beyond binary classification to generate structured eligibility assessments with interpretable reasoning chains that support human-in-the-loop review. This decision support tool represents eligibility as a dynamic state rather than a fixed determination, identifying matches when available and offering actionable recommendations that could render a patient eligible in the future. The system aims to reduce coordinator burden, intelligently broaden the set of trials considered for each patient and guarantee comprehensive auditability of all AI-generated outputs.*

**Introduction**

Applications of artificial intelligence (AI) in healthcare are increasingly focused on improving administrative efficiency and optimizing clinical workflows. One high-value area for these improvements is clinical trial screening[1], where matching patients to appropriate studies remains a major bottleneck. Identifying relevant trials and screening them for a particular patient is traditionally manual, time-consuming, and heavily reliant on clinical expertise. This process is complicated by the heterogeneity of patient data and the complexity of trial protocols. Before eligibility can be assessed, staff must spend considerable time searching vast registries for relevant trials. ClinicalTrials.gov[2], one of the most comprehensive registries, relies on keyword-based search, requiring staff to craft highly specific search terms to locate appropriate trials. These constraints can delay patient access to potentially life-extending treatments and often limit matching efforts to local or regional studies, restricting patient opportunities[1]. Furthermore, insufficient recruitment is the primary reason for trial termination, slowing the overall pace of medical discovery and treatment development[3].

Recent advancements in large language models (LLMs) and agentic AI show growing promise for automating and scaling these tasks, as evidenced by the growing body of literature on the topic[4]. Federal initiatives such as the National Institutes of Health (NIH) TrialGPT project[5] reflect broader momentum toward AI-powered eligibility screening tools. However, tools like TrialGPT suffer shortcomings and there are several areas that stand to benefit from further exploration.

As noted in the TrialGPT discussion, there are a couple limitations. By focusing solely on semantics of inclusion and exclusion criteria, the system does not incorporate structured queries that consider location and trial recruitment status. Additionally, a critical barrier to real-world deployment is the inability to integrate electronic health record (EHR) data. The authors emphasized, "future evaluations should investigate how TrialGPT performs when integrating data from electronic health records…The ability to seamlessly incorporate such diverse data types would significantly enhance the real-world applicability and increase the validation sample size of our framework."[5]:9 Other work in this space is limited by scope and scale[6-8], or implemented in commercial, closed platforms[9-11] that restrict transparency, limit accessibility, and raise concerns about data privacy. Our system is designed with these challenges in mind to deliver a robust, transparent, and secure institutional decision support tool.

Integrating EHR data is challenging because it combines structured fields and unstructured clinical narratives, which often require understanding of domain-specific medical semantics to reliably interpret[12]. Further, successful integration of EHR data requires robust data protections. Compliance with the Health Insurance Portability and Accountability Act (HIPAA) cannot be guaranteed with externally hosted, commercial LLMs (e.g., ChatGPT[13]). Alternatively, open-source models can be locally deployed within compliant, on-premises environments under strict data use controls to meet the security requirements needed to protect clinical data. One such open-source model is DeepSeek-R1[14] ("DeepSeek").

In addition to the advantage of secure deployment, DeepSeek incorporates reasoning-enabled architecture that reflects human-like cognition aligned with real-world clinical workflows[15]. It demonstrates high performance on healthcare-related benchmarks, achieving 92% accuracy on United States Medical Licensing Examination (USMLE) style benchmarks[16], and 93% diagnostic accuracy across 100 diverse MedQA clinical cases[15]. DeepSeek supports a context window of 128,000 tokens, meaning it can process around 100,000 words in a single request, a critical capability for processing extensive documents and particularly important in this application given the length of EHR data and clinical trial protocols. Collectively, these characteristics underscore the model's suitability for interpreting complex medical data and providing decision support in clinical settings.

In this paper, we outline progress toward an AI-aided clinical trial screening system designed to address the limitations of current trial matching methods. Leveraging the reasoning-enabled model, DeepSeek, our system securely ingests EHR inputs and generates eligibility assessments at the patient-trial level, supporting transparency, auditability and flexible implementation. The framework supports real-world use and human-in-the-loop evaluation via clinically informed, templated prompts that maintain logical consistency across evaluations and produce reports that preserve the model's reasoning chains, or steps, in formats familiar to non-technical reviewers.

The system architecture is flexible and can be configured for any secure computing environment. Our system runs on institutionally approved infrastructure, within a National Institute of Standards and Technology (NIST)-compliant data center, enabling secure integration of data and enforcing strict controls for model execution through containerized deployment. For initial experimentation, the system utilizes synthetic data, which facilitates rapid iteration without imposing additional burden on clinical teams. Having demonstrated strong performance on synthetic inputs, the system will advance to the next phase: expert-reviewed validation on a set of real patient cases. The ultimate objective is deployment in real-world healthcare settings.

Effective translational science requires tools that are both cutting edge and accessible, bridging advances in technology with clinical practice. Producing generalizable solutions for common and persistent challenges is a core tenet of The National Center for Advancing Translational Science (NCATS)[17]. By accelerating clinical trial screening, our system addresses patient-trial matching, a challenge faced by clinical coordinators and patients across multiple diseases and conditions. Focusing on deployment feasibility, integration of heterogeneous data sources, and explainability, our system represents progress toward a secure, scalable, and transparent AI tool. Our goal is not to replace clinical expertise but to augment and accelerate coordinator workflows by consolidating the numerous, often siloed, data elements needed for trial screening. By providing clinicians with an accessible and accountable means to review information, the system enables more efficient, consistent, and comprehensive patient-trial matching, ultimately supporting patients with treatment opportunities and facilitating trial enrollment.

**Methods**

We present a proof-of-concept system that combines natural language processing with templated prompting to support flexible data ingestion, interpretable reasoning chains, and integration into real-world clinical workflows. Our system follows a four-step pipeline (Figure 1): (1) a Patient Information Extraction template pulls relevant clinical data from ingested EHR; (2) the output is used to retrieve candidate trials programmatically through ClinicalTrials.gov application programming interface (API)[18]; (3) an eligibility assessment template evaluates patient-trial compatibility using a standardized scoring and reasoning framework; and (4) detailed reasoning for trial eligibility is output with confidence levels and references in multiple formats for human-in-the-loop review.

Initial evaluation of the pipeline uses AI-generated synthetic patient records designed to reflect the variability in clinical documentation. These records include occasional missing fields and heterogeneous formats, such as free text, tables, and PDF-style layouts, which align with real-world presentations of genomic testing results in EHRs[19]. A dedicated parsing module, implemented as a pre-processing step, enhances the system's flexibility in handling these diverse data formats and structures. Using this synthetic dataset allowed us to evaluate whether the system could reliably extract relevant information, generate appropriate search queries to retrieve trials, and produce structured eligibility assessments with interpretable reasoning. This proof-of-concept evaluation demonstrates the generalizability of the core system, establishing a foundation for future tests with real-world data.

A distinctive feature of our pipeline is its use of reasoning models. DeepSeek generates intermediate reasoning steps in `<think>` tags that allow it to work through problems before producing final outputs. This trained reasoning capability, similar to chain-of-thought prompting, helps with complex tasks requiring multi-step problem solving and logical deduction[14]. Through the reasoning process, the model articulates how different pieces of information relate to each other and the problem at hand, creating richer internal context that improves the accuracy of its outputs. By explaining which patient characteristics align with or violate specific trial criteria, the model produces reasoning chains that support clinical validation. Rather than simply accepting or rejecting an AI recommendation based on a final label, clinicians can trace the logical pathway, identify cases where the model may have misinterpreted clinical nuance, and make informed judgments about whether to pursue a given trial match or recommendation. This transparency represents a significant advancement over traditional "black box" AI-aided clinical decision support tools. By leveraging a reasoning-capable model, the system delivers outputs that are not only grounded in relevance, but also increase accountability and foster clinician trust in the system.

In addition to these interpretability benefits, being open source allows us to run the model within local, secure infrastructure rather than interfacing with proprietary services. This combination of rationalized decisions and flexible deployment make DeepSeek an impactful tool as we work toward implementation in healthcare settings.

To standardize interactions with DeepSeek, we utilize the Jinja2 templating language[20] to create reusable prompts for extracting patient information (Step 1 of Figure 1) and assessing trial eligibility (Step 3 of Figure 1). These templates define both the instructions sent to the model, and the expected structured-format responses should follow, ensuring consistent extraction and assessment logic across cases, as well as compatible searching of ClinicalTrials.gov (Step 2 of Figure 1). At runtime, the template is rendered with patient data to dynamically form the prompt passed to the model. By separating the prompt structure (Jinja2 template) from the input data (patient information, trial protocols), the system supports modular and reproducible prompt design. There are two templates in our workflow: the Patient Information Extraction Template, which guides the model from reviewing EHR data to extracting a patient report for downstream processing, and the Patient-Trial Eligibility Evaluator Template, which directs the model in assessing trial-specific inclusion and exclusion criteria, and generating structured, interpretable outputs (Figure 2).

Both templates were informed by subject matter experts experienced in matching patients with clinical trials. Integrating subject matter expertise strengthens the practical utility of AI-augmented clinical decision systems[21,22]. Their feedback on key variables and rule structures was valuable and enhanced the real-world fidelity of our workflow.

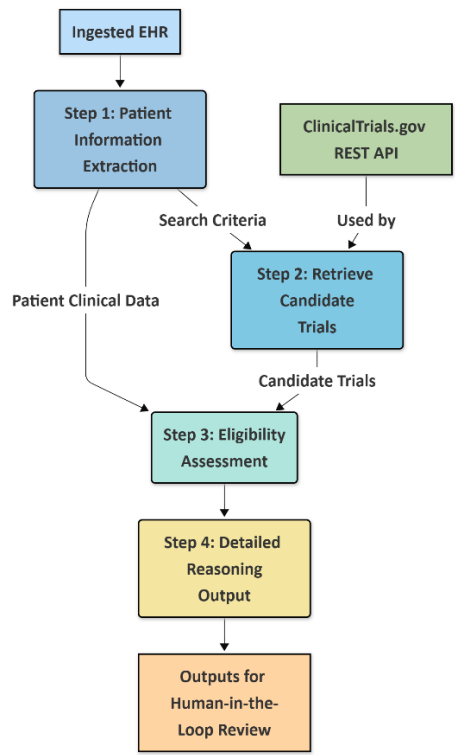

**Figure 1**. Overview of the system pipeline, illustrating data flow and transitions between processing stages.

*1. Patient Information Extraction*

The Patient Information Extraction (PIE) template identifies and extracts 14 key clinical data categories from input EHR data (Table 1). EHR data captures patient-level variables like demographics, diagnoses, prescriptions, treatment notes, laboratory values, and more. Extracted information provides the structured input required for identifying relevant trials and assessing eligibility.

The template provides input category definitions, examples, and extraction rules, which inform the prompt passed to the model to guide its logic. In some cases, rules specify where information is typically located within clinical documents, how to handle medical abbreviations, or the required format for structuring output. For example, the PIE template instructs the model that diagnostic information often appears near the top of documents preceded by terms like "Diagnosis"; it directs the model to extract specific disease type and stage; and if the information is missing or ambiguous, to return "Need more info". Further, extracted diagnosis strings map to the 'condition' field in ClinicalTrials.gov API for trial searches (Step 2 of Figure 1). A loop iterates through each category, processing the definitions, examples, and rules. This programmatic structure allows our system to scale, meaning new categories or rules can be added easily without editing large sections of the prompt. This ensures the same logic is applied across cases.

The PIE template defines the following clinical information categories (Table 1), which are rendered with input EHR data into a prompt sent to the model that informs the structured extraction task.

**Table 1.** Features extracted from patient EHR and the definitions provided in the PIE template.

| Data Element | Description |
|---|---|
| Primary Diagnosis | Disease type, histological subtypes, and staging information used as the primary 'condition' parameter for trial searches. |
| Base Diagnosis | Simplified version of Primary Diagnosis used to broaden trial retrieval. |
| Diagnosis Synonyms | List of synonyms or related keywords derived from Base Diagnosis used in a multi-pass search strategy. |
| Patient Demographics | Age, sex, and geographic location (ZIP Code) formatted for compatibility with API calls to ClinicalTrials.gov. |
| Current Interventions | Active treatments and therapies used as 'intervention' parameters in ClinicalTrials.gov API requests. |
| Treatment History | Comprehensive prior therapy documentation including dates, responses, and outcomes. |
| Search Terms | Broader disease categories and staging information for expanded trial matching. |
| Biomarkers / Molecular Profile | Genetic testing results, protein expression levels, and molecular characteristics. |
| Performance Status | Functional status indicators (e.g., Eastern Cooperative Oncology Group (ECOG) status, Karnofsky Performance (KP)scales). |
| Laboratory Values | Hematologic parameters and organ function markers with units and any referenced flags. |
| Comorbidities | Co-existing medical conditions that may impact trial eligibility. |
| Family History | Relevant familial medical history and genetic predisposition factors. |
| Treatment Goals | Therapeutic objectives and care planning information. |
| Eligibility Factors | Specific factors impacting clinical trial participation (e.g., smoking history, organ function adequacy). |

*2. ClinicalTrials.gov API Call*

One hurdle in system development required optimizing the specificity of search terms used for trial retrieval. In early stages of development, the LLM extracted highly specific primary diagnoses that, when passed through the ClinicalTrial.gov API, returned limited results. The issue was not in the diagnosis extracted, these terms correctly reflected the precise clinical language in the input data, rather there was variability in how equivalent conditions were indexed (e.g., "adenocarcinoma of the bronchus" vs. "non-small cell lung cancer"). ClinicalTrials.gov lacks semantic similarity search functionality, meaning equivalent clinical terms are not recognized. If server-side similarity matching were available, these differences could be addressed more systematically.

In the absence of such features, we employ custom logic using a multi-tiered keyword search strategy that balances specificity with comprehensive trial retrieval. The system performs multiple searches with descending levels of specificity: first using the precise primary extracted diagnosis, then using broader diagnosis terminologies which we call "Diagnosis Synonyms." This approach reflects a deliberate decision to cast a wider net in the retrieval phase. Since an AI system, rather than a human screener, conducts the initial filtering, we prioritized recall over precision making our system more tolerant of false positives (retrieving potentially irrelevant trials) while avoiding false negatives (missing relevant trials). This results in a larger pool of candidate trials for subsequent evaluation.

It was important to balance search breadth with computational load. Very broad searches of the database could return thousands of trials, which increases storage requirements and the compute costs of downstream evaluation. Considering this, we defined a set of minimum search criteria to narrow candidate trials without compromising coverage. Minimum search criteria include recruiting status, only retrieving trials that are actively recruiting participants, as well as demographics, like age and gender considerations that might exclude participation outright. Geographic location is also considered, with a default restriction to trials within the United States. Together, these fields form the base of our multi-pass search strategy.

For each search, our system constructs a set of query parameters designed to maximize trial coverage. The minimum search criteria are supplied with each query to iteratively retrieve trials. The system consolidates results from multiple queries, removes duplicates, and stores each candidate trial in its own structured JavaScript Object Notation (JSON) object capturing both metadata (`NCT_ID`, and `Trial_Title`) and protocol details (`description`, `inclusion_criteria`, `exclusion_criteria`). This approach allows the downstream eligibility evaluation assessment to reference a comprehensive and curated set of trials for each patient. This search set-up ensures the resulting trial set is both broadly relevant and tailored to basic patient-specific characteristics. Summary statistics, including the total number of retrieved trials, are recorded to characterize dataset composition. When running the pipeline on 30 synthetic patient EHRs, the average number of candidate trials identified was 950 trials per patient, representing a tenfold increase prior to incorporating synonym-based expansion.

*3. Patient-Trial Eligibility Assessment*

Following data extraction, the consolidated patient report generated in Step 1 and the associated bundle of candidate trials identified in Step 2 are evaluated using our Patient-Trial Eligibility Evaluator (PTEE) template, the third step in the pipeline (Figure 1). This assessment step is executed within the overarching `run_pipeline` script. The PTEE template is rendered with structured patient-specific information across 14 categories (Table 1) and individual trial protocols to generate contextually relevant prompts for trial-by-trial compatibility analyses.

Like the PIE template, the PTEE template was developed in collaboration with subject matter experts. It provides a structured assessment framework that prompts the model to consider specific aspects of eligibility, from basic demographics to complex clinical criteria. The PIEE template operationalizes a three-part assessment structure encompassing (1) minimum eligibility criteria, (2) inclusion criteria, and (3) exclusion criteria.

First, the templated prompt instructs the model to verify demographic criteria are satisfied and to reference the supporting data points. Next, the model assesses the broader clinical profile, analyzing inclusion and exclusion criteria separately to maintain clarity. Patient data are compared against trial protocol specifications. For each determination, the PIEE template instructs the model to provide its reasoning and indicate its level of certainty based on the clarity and completeness of data. Every evaluation decision includes an evidence-based explanation referencing relevant patient and protocol information, highlighting gaps in knowledge due to missing or ambiguous information, and offering recommendations for follow-up testing or needed clinical clarification. These outputs are aggregated and the model produces an overall recommendation for the patient-trial pair:

- *Eligible Now*: Patient fully meets all key eligibility criteria.

- *Could Be Eligible in Future*: Patient does not currently meet all criteria but may become eligible later.
- *Not Eligible*: Patient does not meet key eligibility criteria.
- *Need More Information*: Insufficient data to determine eligibility.

This approach transforms the human-in-the-loop reviewer's task from exhaustively checking every assessment to focusing their expert judgment where it matters most: verifying matches with high certainty and resolving ambiguous cases.

Evaluating eligibility is computationally intensive because each patient must be assessed against every candidate trial. To reduce runtime, the system supports multi-threaded execution, allowing concurrent patient-trial evaluations to be processed in parallel. Multi-threading is particularly effective in LLM workflows because most of the latency comes from model inference[23], so performing inference in batches with multiple chat completions generating at once results in a higher throughput.

*4. Detailed Reasoning Output for Expert Review*

A key consideration in developing our clinical trial matching system was the accessibility for non-technical users. System outputs are returned as JSON files (Figure 2 shows an example) and programmatically formatted via a Python script into Word and PDF reports, preserving the model's reasoning chains and presenting results in a hierarchical structure. Overall assessment decisions are summarized upfront with supporting evidence and reasoning provided in subsequent sections. This formatting aligns outputs with familiar documentation practices, supporting efficient human review. Expert oversight is crucial in clinical AI deployments, particularly in clinical trial coordination with patients[22], and is therefore a central component of the system. Each patient-trial report includes assessment metadata that contains temporal information like assessment date, along with key identifiers including `trial_ID`, `patient_ID`, and `assessor_information` (i.e., which model was used, which version of a template was used) for auditing purposes. This approach prioritizes workflow interpretability, preserving transparency from input through assessment to support evidence-based decision making.

**Discussion**

We evaluate our pipeline with synthetic electronic health records simulating diverse clinical presentations. Patient-trial eligibility assessments are aggregated in a post-processing step to obtain global performance metrics and to examine how the system reasons through eligibility decisions at the patient level across a range of clinical scenarios. These analyses demonstrate that the system functions as intended, providing a strong foundation for future work.

The majority of assessments were classified as *Not Eligible* (25,192 of 28,575 total patient-trial evaluations, or 88%). This distribution aligns with clinical expectations; identifying trials that match unique patient presentations is inherently challenging and there are known imbalances in the clinical trial landscape. Conditions like breast, lung, and hematologic cancers (e.g., lymphoma and leukemia) make up a significant proportion of trial volume compared to other, rarer cancer types such as pancreatic, which are substantially less represented[24,25]. In one case, when processing a synthetic patient with Hodgkin lymphoma, the system evaluated over 5,000 eligible trials and identified eight matches. In contrast, when processing another synthetic patient with adenoid cystic carcinoma of salivary gland, a rarer cancer[26], the system reviewed nearly 800 trials, six-fold fewer, and found only one match. This illustrates how condition rarity directly reduces trial opportunities and constrains available matches, simultaneously demonstrating the system's ability to scale appropriately with available evidence.

Only a small fraction of assessments are classified as *Eligible Now*, with 16 of the 30 patient personas receiving at least one trial match. This limited number of matches reflects the conservative design of the evaluation template, which prompts the system to explicitly reason through inclusion and exclusion criteria and avoid proposing eligibility without sufficient justification. Among patients who did not receive *Eligible Now* determinations, the system frequently produced *Need More Information* or *Could Be Eligible in the Future* classifications. These labels illustrate the reasoning model's capacity for temporal and protocol-grounded assessment, with explanations typically reflecting missing clinical data or criteria that could change over time. This indicates that enrollment opportunities could be within reach through additional tests, imaging, or therapy interventions. The model's reasoning offers actionable insights that can guide clinical decision-making.

Take the example below, a synthetic patient with pancreatic adenocarcinoma is evaluated against trial NCT05764720[27], an excerpt from the corresponding evaluation report is shown below (Figure 2):

```json
{
    ... // Assessment metadata omitted for brevity
    },
    "eligibility_summary": {
        "eligibility_status": "Could Be Eligible in Future",
        "confidence_level": "MEDIUM"
        ... // Summarized assessment omitted for brevity
    },
    "primary_criteria_assessment": [
        ... // Age, sex, and geographic assessments omitted for brevity
        {
            "name": "Primary Diagnosis Match",
            "status": "Eligible Now",
            "reasoning": "Borderline resectable pancreatic adenocarcinoma (T2N1M0 Stage IIB) matches trial focus"
        }
    ],
    "clinical_criteria_assessment": [
        {
            "name": "Biomarker/Molecular Profile",
            "status": "Eligible Now",
            "reasoning": "KRAS G12D mutation present (common in PDAC), trial doesn't specify biomarker requirements"
        },
        {
            "name": "Prior Therapy Compatibility",
            "status": "Could Be Eligible in Future",
            "reasoning": "Requires ≥2 months chemotherapy - patient has only completed 2 weeks of FOLFIRINOX"
        }
        ... // 3 additional sub-assessments omitted for brevity
    ],
    "exclusion_criteria_assessment": [
        ... // 4 other records removed for brevity
        {
            "criterion": "Gross tumor invasion of stomach/duodenum",
            "status": "Need More Information",
            "reasoning": "Imaging details not provided in patient data"
        }
    ],
    "actionable_recommendations": [
        "Complete 2 months of FOLFIRINOX chemotherapy (currently at cycle 1/12)",
        "Verify ability to interrupt systemic therapy for radiation per protocol requirements",
        "Obtain imaging confirmation of tumor anatomy/adjacent organ involvement",
        "Test breath-hold capability (≥20 seconds required)"
        ... // 2 more recommendations omitted for brevity
    ],
    "missing_data_points": [
        "Geographic location/distance to trial site",
        "Imaging confirmation of tumor anatomy",
        "Breath-hold capability assessment"
    ]
    ... // qa checklist records and final closing brace omitted for brevity
}
```

**Figure 2.** A stylized example of the JSON produced by the reasoning model capturing reasoning. Portions are truncated for brevity.

In this example, the extracted patient report, result of Step 1, states the patient completed only one two-week cycle of FOLFIRINOX, a chemotherapy regimen, but the trial requires at least 2 months of lead-in chemotherapy. Rather than classifying this patient as *Not Eligible*, the system correctly recognizes this as a deterministic temporal prerequisite and assigns a *Could Be Eligible in Future* label with medium certainty. The model's reasoning chain distinguished this time-dependent requirement from other unresolved factors, including breath-hold capability, therapy interruption feasibility, and confirmatory imaging. These clinical assessments could be completed and result in potentially decisive outcomes which justifies the model's medium confidence designation. While the chemotherapy duration requirement is straightforward to satisfy with time, outstanding clinical uncertainties remain determinative. If any of these factors fall outside the trial's preferred values, the patient would become ineligible regardless of time on their current intervention.

Each actionable recommendation generated maps to explicit protocol criteria, demonstrating that the model represents eligibility as a dynamic state rather than a static binary classification. This temporal reasoning capability prevents premature exclusion of potentially eligible candidate trials while maintaining appropriate clinical caution. Furthermore, the transparent reasoning chain supports clinical coordinators in their review, offering insights into determination and suggesting specific next steps for advancing patients toward eligibility.

The modular, template-driven architecture of our system offers flexibility for customization across clinical contexts. The PIE template in Step 1 can be easily modified to capture additional data elements without requiring detailed intervention from the technical team. Clinicians can suggest condition-specific edits that would automatically matriculate through the system during processing. This configurability extends to trial retrieval in Step 2. While our current implementation searches the entire ClinicalTrials.gov registry, the pipeline can be adapted to assess specific pre-identified trials if a coordinator already has selected trials of interest. Similarly, the PTEE template in Step 3 can be customized to incorporate specific reasoning steps, or to adopt more comprehensive or permissive assessment modes. Such targeted deployment scenarios streamline the manual review process. Overall, this flexible architecture positions the system to support diverse use cases as clinical needs shift.

Human-in-the-loop review is both expected and critical to the system's overall success. An important area identified for future development is the integration of source citation capabilities. To support this, we plan to build out the system's ability to explicitly reference the precise source location of extracted data elements (i.e., report date, page number). This traceability directly supports our guiding principles of transparency and auditability. Another priority for future development is the creation of an intuitive web-based interface. While the current system supports expert review through deliberate formatting and file exports, a dedicated interface would facilitate expert validation and support system improvements. Systematic capture of expert review data could be leveraged through reinforcement learning to improve the system's performance over time, establishing an iterative refinement loop grounded in real-world clinical expertise. Following further refinement and evaluation with real-world data, we plan to release the generalized pipeline on GitHub, an open platform promoting accessibility and community-driven improvements.

**Conclusion**

This work presents a configurable, transparent, and reasoning-capable clinical trial matching system that advances existing methodologies through the integration of reasoning-enabled LLMs and dynamic templated prompts. A distinguishing feature of reasoning models is their ability to generate detailed rationales alongside responses, a functionality particularly important in clinical settings where validation of AI outputs is paramount. The use of dynamic templating through the Jinja2 language improves prompting efficiency and facilitates customization. These elements embody our guiding principles of transparency, auditability, and flexible implementation, qualities essential for real-world deployment.

The system addresses two fundamental challenges in clinical trial recruitment: scale and complexity. It supports large-scale trial evaluations across diverse clinical scenarios and prioritizes human-in-the-loop verification as a principal step in the workflow. The comprehensive clinical trial screening pipeline extracts relevant patient information from input clinical documentation, standardizes these data for compatibility with ClinicalTrials.gov registry searches, and conducts patient-trial eligibility assessments using reasoning-enabled evaluation. Outputs include structured reports with supporting justifications and confidence indicators, providing clinicians with interpretable and actionable decision support for informed trial matching. By augmenting the labor-intensive process of matching patients with appropriate trials, the system addresses a significant bottleneck in clinical trial recruitment. It offers clinicians an accessible and accountable AI tool for comprehensive patient-trial screening, ultimately connecting more patients to potentially life-extending investigational opportunities and facilitating trial enrollment. By supporting both patients and clinicians in the task of clinical trial matching, the system accelerates enrollment, reduces burden, and strengthens the efficiency and accessibility of clinical research.

**Acknowledgements**
This project was supported by the NIH National Center for Advancing Translational Sciences through grant number UL1TR001998, and the University of Kentucky Markey Cancer Center (MCC) through grant number P30CA177558. The content is solely the responsibility of the authors and does not necessarily represent the official views of the NIH or the MCC.